\documentclass{egpubl_arxiv}
\usepackage{eg2024}

\usepackage[T1]{fontenc}
\usepackage{dfadobe}  

\biberVersion
\BibtexOrBiblatex
\usepackage[backend=biber,bibstyle=EG,citestyle=alphabetic,backref=true]{biblatex} 
\electronicVersion
\PrintedOrElectronic

\ifpdf \usepackage[pdftex]{graphicx} \pdfcompresslevel=9
\else \usepackage[dvips]{graphicx} \fi

\usepackage{egweblnk} 

\usepackage{amsmath,amsfonts,bm}

\def\source{\mathcal{S}}
\def\target{\mathcal{T}}
\def\domain{\mathcal{Q}}
\def\motion{M}
\def\pose{P}
\def\StoT{\mathcal{G}}
\def\TtoS{\mathcal{F}}
\def\mmotion{\Tilde{\motion}}
\def\mpose{\Tilde{\pose}}
\def\D{\mathcal{D}}
\newcommand{\Loss}[1]{\mathcal{L}_{\textrm{#1}}}
\def\fid{n}
\def\nframe{N}
\def\poseSeq{\left[ \pose_{\fid} \right]_{\fid=1}^{\nframe}}
\def\rootSeq{\left[ R_{\fid} \right]_{\fid=1}^{\nframe}}
\def\mposeSeq{\left[ \Tilde{\pose}_{\fid}^{\target} \right]_{\fid=1}^{\nframe}}
\def\mrootSeq{\left[ \Tilde{R}_{\fid}^{\target} \right]_{\fid=1}^{\nframe}}
\def\footJ{\Phi}
\def\eeJ{\Theta}

\def\MS{\motion^{\source}}

\def\PS{\pose^{\source}}
\def\PT{\pose^{\target}}
\def\restP{\pose_0}

\def\Tableref#1{Table~\ref{#1}}
\def\figref#1{fig.~\ref{#1}}
\def\Figref#1{Fig.~\ref{#1}}

\def\eqref#1{eq.~\ref{#1}}

\def\1{\bm{1}}

\DeclareMathAlphabet{\mathsfit}{\encodingdefault}{\sfdefault}{m}{sl}
\SetMathAlphabet{\mathsfit}{bold}{\encodingdefault}{\sfdefault}{bx}{n}

\newcommand{\Ls}{\mathcal{L}}
\newcommand{\R}{\mathbb{R}}

\DeclareMathOperator*{\argmax}{arg\,max}

\usepackage{booktabs} 
\usepackage{paralist}
\usepackage{makecell}
\usepackage{multirow}
\usepackage{array}
\newcolumntype{L}[1]{>{\raggedright\let\newline\\\arraybackslash\hspace{0pt}}m{#1}}
\newcolumntype{C}[1]{>{\centering\let\newline\\\arraybackslash\hspace{0pt}}m{#1}}
\newcolumntype{R}[1]{>{\raggedleft\let\newline\\\arraybackslash\hspace{0pt}}m{#1}}

\usepackage[font=small,skip=2pt]{caption}

\usepackage[ruled]{algorithm2e} 
\usepackage{cleveref}
\Crefformat{figure}{#2Fig.~#1#3}
\Crefmultiformat{figure}{Figs.~#2#1#3}{ and~#2#1#3}{, #2#1#3}{ and~#2#1#3}

\SetAlFnt{\small}
\SetAlCapFnt{\small}
\SetAlCapNameFnt{\small}
\SetAlCapHSkip{0pt}

\newcommand{\qq}[1]{{{{#1}}}}

\usepackage{bbold}

\usepackage{xpunctuate}
\providecommand{\eg}[0]{e.g\xperiod}
\providecommand{\ie}[0]{i.e\xperiod}
\providecommand{\etc}[0]{e.t.c\xperiod}
\providecommand{\st}[0]{s.t\xperiod}

\newcommand{\ours}{Pose-to-Motion\xspace}

\title[Pose-to-Motion]%
      {Pose-to-Motion: Cross-Domain Motion Retargeting with Pose Prior}

\author[Q. Zhao \textit{et al.}]
{\parbox{\textwidth}
{\centering Qingqing Zhao$^{1}$~~Peizhuo Li$^{2}$~~Wang Yifan$^{1}$~~Olga Sorkine-Hornung$^{2}$~~Gordon Wetzstein$^{1}$
}
\\
{\parbox{\textwidth}{
\centering $^1$Stanford University~~$^2$ETH Zurich
}
}
} 

\begin{document}

\teaser{
 \vspace{-35pt}
 \includegraphics[width=0.95\textwidth]{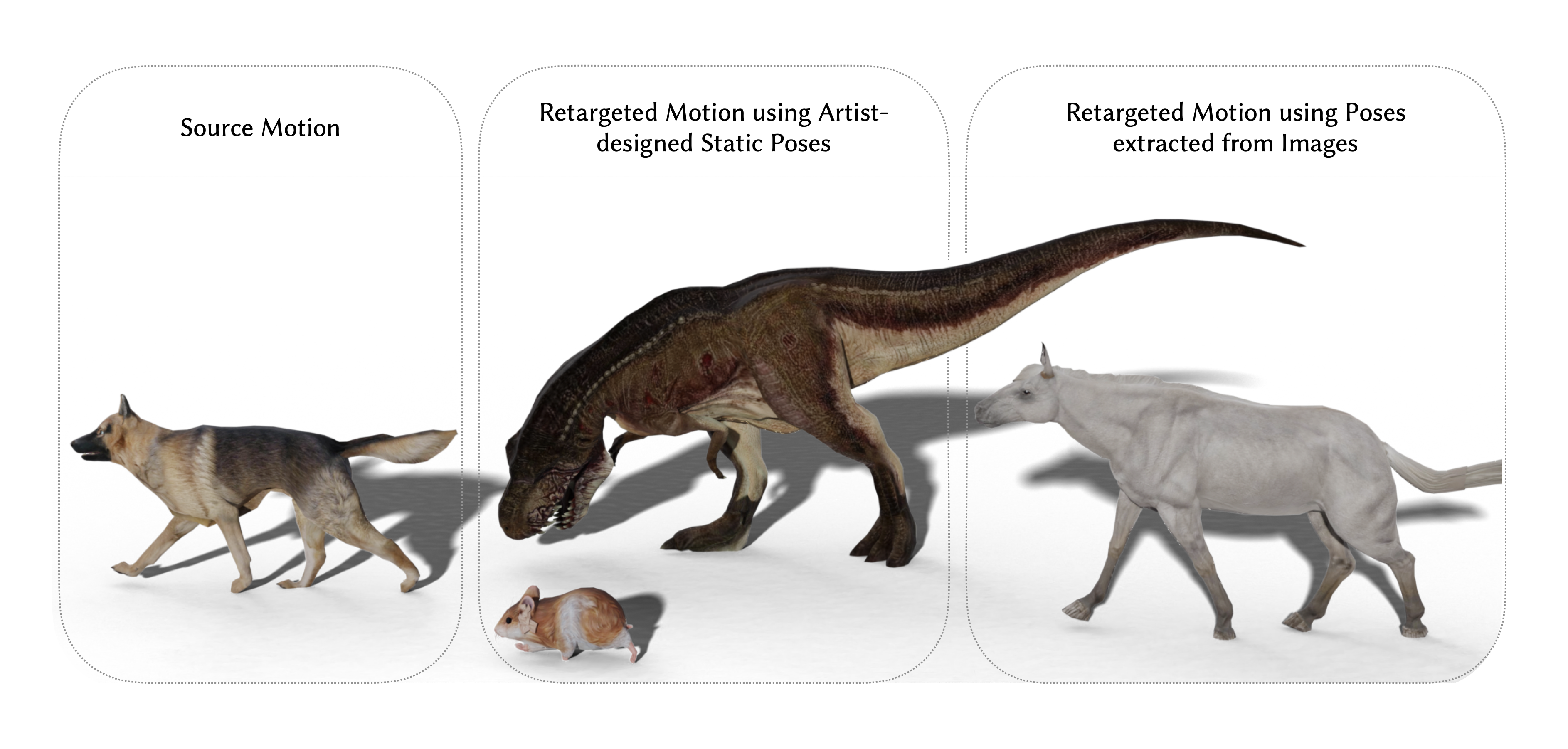}\centering
  \caption{Our generative motion retargeting framework enables the motion dynamics of one creature to be transferred to another in a plausible manner. For this purpose, motion capture data of the source creature is transferred to the target while taking a few static body poses of the target into account, among other constraints. This allows us to transfer the motion dynamics of tame and cooperative animals, such as dogs, to more exotic creatures, such as horses, rodents, or carnivorous dinosaurs, for which motion capture data may be difficult to obtain but individual body poses are readily available.  
  }
  \vspace{10pt}
  \label{fig:teaser}
}

\maketitle
\begin{abstract}
Creating believable motions for various characters has long been a goal in computer graphics.
Current learning-based motion synthesis methods depend on extensive motion datasets, which are often challenging, if not impossible, to obtain. 
On the other hand, pose data is more accessible, since static posed characters are easier to create and can even be extracted from images using recent advancements in computer vision.
In this paper, we utilize this alternative data source and introduce a neural motion synthesis approach through retargeting. Our method generates plausible motions for characters that have only pose data by transferring motion from an existing motion capture dataset of another character, which can have drastically different skeletons.
Our experiments show that our method effectively combines the motion features of the source character with the pose features of the target character, and performs robustly with small or noisy pose data sets, ranging from a few artist-created poses to noisy poses estimated directly from images. Additionally, a conducted user study indicated that a majority of participants found our retargeted motion to be more enjoyable to watch, more lifelike in appearance, and exhibiting fewer artifacts. The project page can be found here: \url{https://cyanzhao42.github.io/pose2motion}.


\end{abstract}


\section{Introduction}

The ability to generate plausible motion across a diverse array of characters is a crucial aspect of creating immersive and engaging experiences in this digital era, and is vital to a wide range of applications including augmented reality, cinematography, and education.
Recently, motion retargeting from unpaired motion data has emerged as a promising approach to address these needs~\cite{gao2018automaticUS,villegas2018neural, aberman2020skeleton,villegas2021contact,zhang2023skinned}.
However, these techniques depend heavily on high-quality motion data, which can be difficult to acquire despite notable progress in 4D video-based reconstruction, as those methods paradoxically depend on substantial amounts of meticulously annotated videos that are often impossible to collect for unique non-humanoid characters.
On the other hand, recent advancements in computer vision have enabled unsupervised 3D pose extraction from single images, providing a more accessible data source \cite{wu2023magicpony}. 
Our work seeks to tap into this alternative data source for motion retargeting to enable motion synthesis for a variety of domains where only pose data is available, thereby broadening the application of motion retargeting for motion synthesis in domains where MoCap data is scarce. 

Reducing data requirements has always been a key goal in motion retargeting research.
Powered by cycle-consistent generative adversarial networks~\cite{zhu2017unpaired, goodfellow2014generative}, recent approaches have moved away from traditional approaches, which either require skeleton-level correspondence~\cite{gleicher1998retargetting,choi2000online,monzani2000using,popovic1999physically,tak2005physically,villegas2021contact,zhang2023skinned} or pose-level correspondence~\cite{sumner2004deformation,baran2009semanticDT,seol2013creatureFO,wampler2014generalizing,celikcan2015exampleBasedRO,abdulmassih2017motionSR}, successfully demonstrating the capability to transfer motion between different skeletal structures using unpaired motion data~\cite{villegas2018neural,aberman2020skeleton,gao2018automaticUS,Adult2child}, as long as they are topologically similar.
However, these approaches largely depend on having symmetric data, \ie, a similar amount of high-quality motion data from both the source and target domains is required, which can be challenging and sometimes impractical to obtain.

In this paper, we propose a novel approach, \ours{}, which leverages pose data from the target domain - which is more accessible than motion data, for example by artist creation, by estimation from images using contemporary computer vision techniques, or by analyzing the fossils of extinct creatures,
to tackle the fundamental challenge posed by the scarcity of high-quality motion data.
Our method leverages a unique \emph{asymmetric} CycleGAN, transforming source domain motion data to target domain pose data and vice versa, effectively allowing us to ``project'' motion onto our target characters using solely their pose data (\Figref{fig:overview}).
This cycle is further refined by synthesizing plausible root transformations using soft constraints, overcoming the root ambiguity problem arising from the lack of motion data in the target domain.
While neither cycle consistency nor adapted soft constraints are novel concepts, applying them to asymmetric data within the realm of motion retargeting offers a new and effective solution to the unique challenges we face in our task. This approach specifically addresses the large domain gap between motion clips in the source domain and static poses in the target domain.
As we demonstrate in \cref{sec:experiments},
our method is able to generate plausible motion for a wide range of subjects by combining the motion prior from another domain, where motion capture (MoCap) data has been captured a-priori, and the pose prior of the subject observed from static poses, even when the pose data is small or noisy.


\ours{} introduces a new perspective in the field of motion synthesis, suggesting a method to produce motion data for a variety of characters without relying heavily on extensive, character-specific motion capture sessions.
In summary, this paper makes the following primary contributions:
\begin{compactenum}[1.]
\item We propose a novel motion-retargeting approach for motion synthesis, \ours{}, which leverages pose data from the target domain to tackle the fundamental challenge posed by the scarcity of high-quality motion data.
\item We demonstrate how this approach leverages asymmetric cycle consistency and soft constraints to synthesize root transformations, overcoming the unique root ambiguity issue in the pose-to-motion setting.
\item We present a detailed analysis and comparison of our method against existing motion-to-motion and pose-to-pose retargeting approaches, showing state-of-the-art results in terms of motion quality and versatility across a wide range of characters and poses.
\end{compactenum}

\section{Related work}


\begin{figure}
    \centering
    \includegraphics[width=\linewidth, clip, trim=0 0 0 0cm]{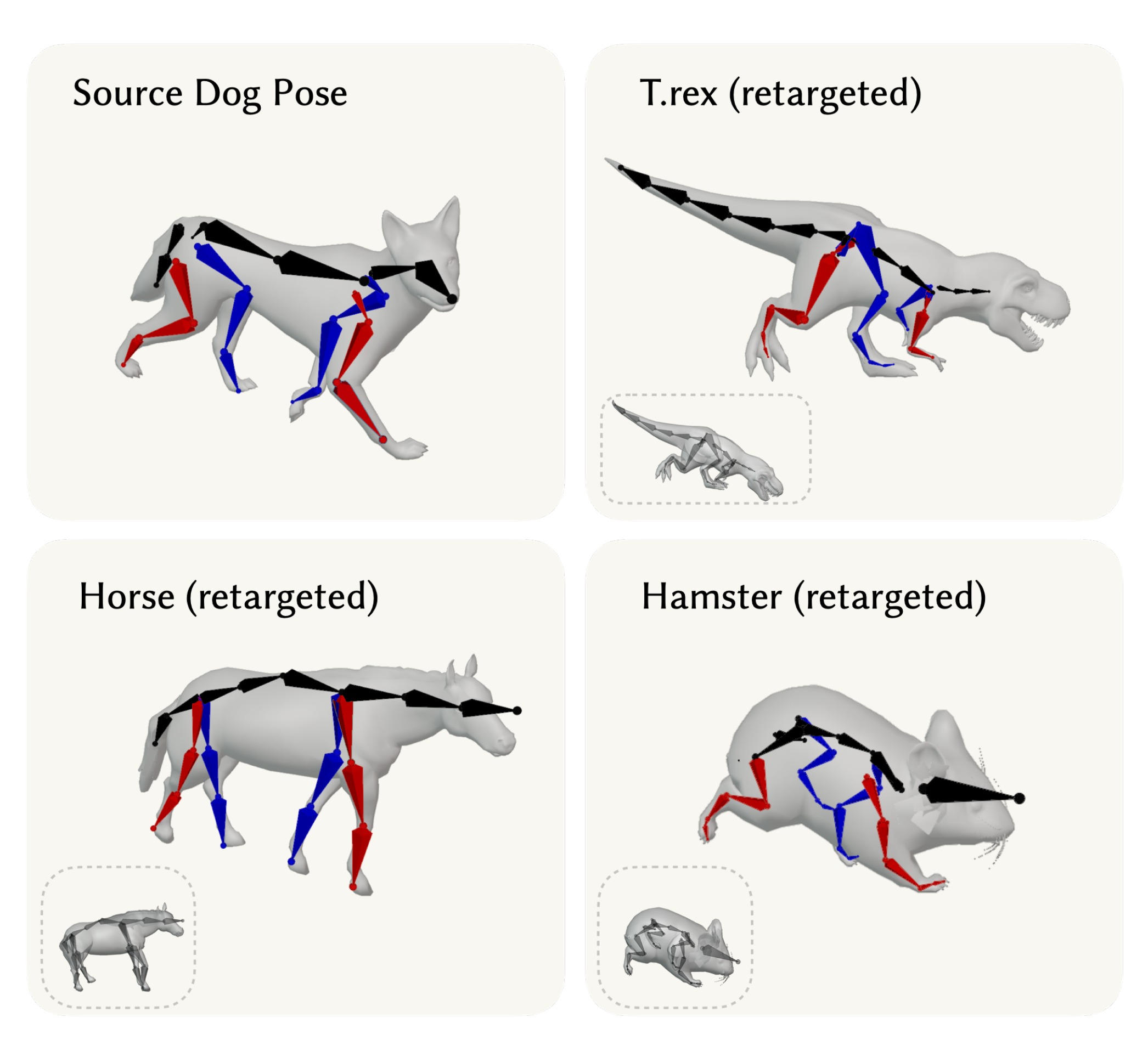}
    \caption{\textbf{Motion retargeting to versatile characters.} Given a small set of artist-created posed animals (\eg t.rex, hamster) or noisy poses derived from 2D images (horse), our method successfully transfers the dog motion to these animals despite significant differences in their bone structures. We include images of the closest instance in the training data at the lower left corner, highlighting the preservation of key attributes during the motion retargeting process. Notably, the elongated tail of the T.rex, the arched spine of the hamster, and the forward-bending knee of horse are all preserved even though the source dog pose does not contain these characteristics. Please refer to the supplementary video for additional qualitative evaluation of the motion clips.}
    \label{fig:zoo_variety}
\end{figure}

\paragraph*{Motion Retargeting.}
As one of the pioneering works, \cite{gleicher1998retargetting} proposed to solve the kinematics constraint of two topologically identical skeletons with a space-time optimization problem. \cite{lee1999hierarchical,choi2000online} further employ per-frame inverse kinematics (IK) for retargeting, followed by a smooth process while preserving high-frequency details. \cite{monzani2000using} explore the possibility of using an intermediate skeleton to retarget motion between skeletons with different numbers of bones. In addition to simple kinematics constraints, \cite{popovic1999physically} introduce dynamics constraints to the spacetime optimization and achieves better realism of the source motion sequence. \cite{tak2005physically} take a different approach by modeling the retargeting problem as a state estimation on a per-frame Kalman filter and further improve the realism of generated motion.

However, those methods are limited to retargeting between motions with skeletons containing limited differences in bone proportion, thus are unable to handle retargeting between different creatures. Since the desired motion gaits and the correspondence between motions cannot be inferred solely from bone proportions in these cases, especially for drastically different creatures like humanoid and quadrupeds, \cite{baran2009semanticDT} propose to exploit a few sparse mesh pairs to transfer poses between different creatures using feature extraction and extrapolation. \cite{yamane2010animatingNC} also exploits paired poses and is able to retarget motion to a different creature. \cite{seol2013creatureFO} demonstrate the ability to control a target creature with human motion, given paired motion examples. \cite{wampler2014generalizing} model the locomotion of different creatures with the same skeletal structure with a physically-based optimization method, and is able to capture the gait with a few examples. However, it requires delicate handcrafted design and is limited to locomotion. \qq{\cite{ikemoto2009generalizing} utilizes Gaussian processing and probabilistic inference to map motion from one control character to a different target character, but it requires artists' edits as training data.}

\cite{celikcan2015exampleBasedRO} use paired pose and mesh to retarget motion from humans to different meshes. Although they are able to perform retargeting between different creatures, at least several paired poses or motions are required as guidance. Besides, when applying pose transfer method to motion in a frame-by-frame manner, the high-frequency details of motion and the temporal coherence are not well preserved, and the missing global translation information leads to severe foot skating artifacts. The same issue also applies to the generative model for poses~\cite{SMPL-X:2019} learned from a large dataset, making motion generation with only pose prior extremely challenging. An interesting exception is the work of \cite{abdulmassih2017motionSR}, which requires only a manually assigned part correspondence to achieve motion style transferring between different creatures.

\paragraph*{Neural Motion Processing.}
With the progress of deep learning, deep neural networks are applied to motion process and synthesis tasks \cite{aberman2020unpaired,yin2023dance,li2022ganimator,shimada2020physcap,li2021ai}, including recurrent neural networks (RNNs)~\cite{fragkiadaki2015recurrent,andreou2022pose}, convolutional neural networks (CNNs)~\cite{holden2015learning,holden2016deep}. As for motion retargeting, \cite{jang2018variational} apply a U-Net structure to paired motion data to solve the problem. \cite{villegas2018neural} uses cycle-consistency adversarial training~\cite{zhu2017unpaired} on a RNN for retargeting, and drops the requirement for paired motion datasets. \qq{\cite{Adult2child} uses cycle-consistency training to transform adult motion capture data to the style of child motion, trained on a small number of sequences of unpaired motions from both domains.} PMnet~\cite{lim2019pmnet} opt for CNNs and achieve better performance. With the proposed skeleton-aware networks, \cite{aberman2020skeleton} can retarget among skeletons with different yet homeomorphic topologies. \cite{li2022iterative} bypass the usage of adversarial training and use an iterative solution with a motion autoencoder.  At the same time, directly transferring poses without any correspondence information pose-wise and geometry-wise is made possible with neural networks~\cite{gao2018automaticUS,liao2022skeleton}. More recent works keep exploring the possibility of better retargeting results by incorporating skinning constraints introduced by the geometry~\cite{villegas2021contact, zhang2023skinned}. Note those methods require motion dataset on both source and target skeleton for training, but the difficulty of acquiring high-quality and comprehensive motion dataset greatly limits their usage. We demonstrate that we can achieve similar performance as skeleton-networks~\cite{aberman2020skeleton} on the Mixamo dataset~\cite{mixamo} in~\Cref{sec:mixamo}, while our model is trained only with a pose dataset for the target character.


\section{Method}
\begin{figure*}[htbp]
    \centering
    \includegraphics[width=0.9\linewidth]{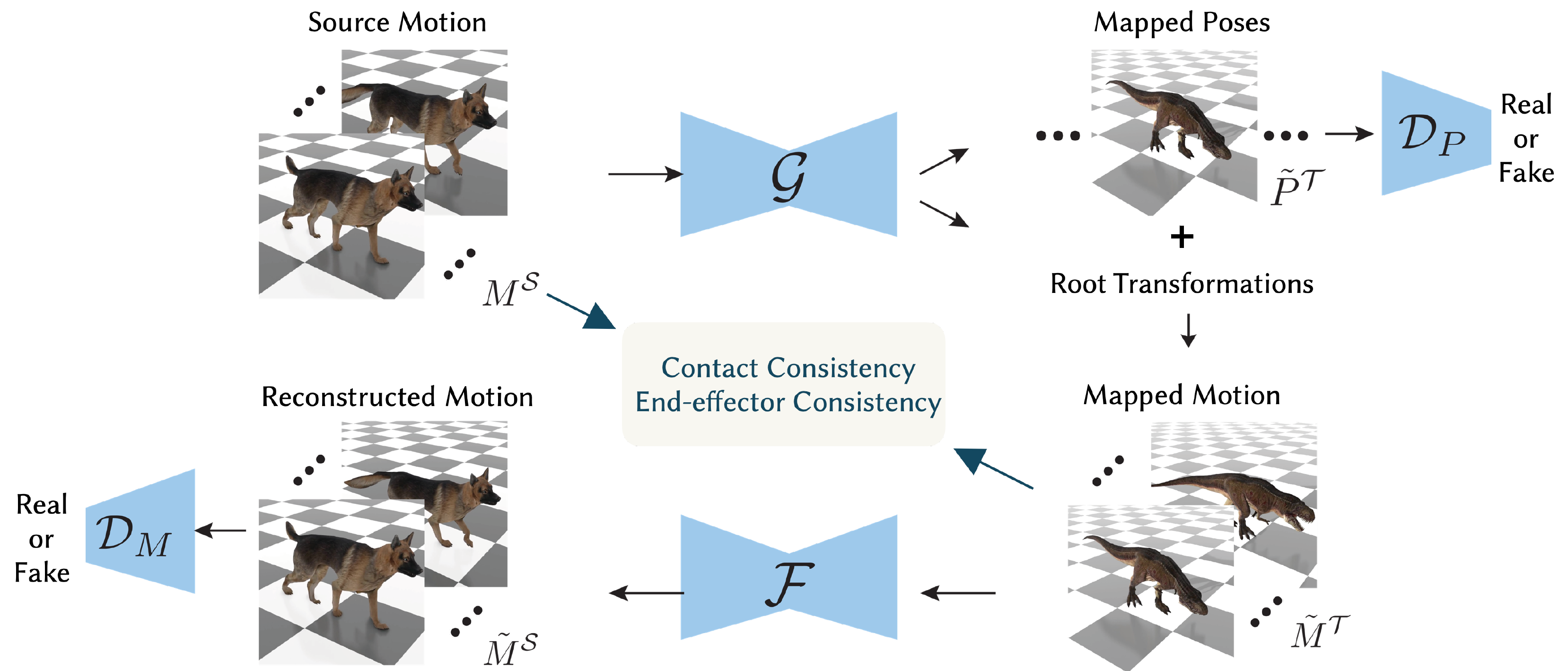}
    \vspace{20pt}
    \caption{\textbf{Method overview.} Our method builds on an asymmetric CycleGAN. The first half of the cycle maps a motion sequence from the source domain \(\MS\) to a sequence of poses \(\mposeSeq\) and root transformations \(\rootSeq\)in the target domain; the individual poses are compared against the pose dataset of the target domain using a pose discriminator \(\D_{\pose}\). The other half of the cycle maps the sequence of poses and root transformations (\(\mmotion^{\target}\)) back to a motion sequence in the source domain \(\mmotion^{\source}\), which is supervised with a reconstruction loss and an adversarial loss using a motion discriminator. The contact and end-effector consistency implicitly regulates the root prediction, leading to more realistic motion.}
    \label{fig:overview}
\end{figure*}

\subsection{Data representation.}\label{sec:representation}
We inherit the representation for pose and motion from prior work \cite{aberman2020skeleton}, which we briefly recap below.
Given the skeleton structure of a character with \(J\) joints, its pose is represented using a vector $\pose \in \R^{6J}$, defining the relative joint rotations in the kinematic tree, where each rotation is represented by a \(6\)-dimentional rotation \cite{zhou2019continuity}.

A character's motion consists of a sequence of poses \(\poseSeq\) and root transformations \(\rootSeq\), where \(R_{\fid}\) is composed of root orientation \(\theta_{r}\in\R^{6}\) and velocity \(v_{r}\in\R^{3}\). The root transformation is handled as a special armature connected to the root node. As such, the overall representation of motion can be denoted by $\motion\in\R^{T\times \left(6\left(J+1\right)+3\right)}$, where \(T\) is the number of frames in the sequence.
Note that the root displacement, $x_r$, can be computed from root velocity using the forward Euler method \(x_r(t+1) = x_r(t)+v_r(t)\).


In the following, we denote the motion and pose from a domain \(\domain\in\left\{ \source, \target \right\}\) as $\motion^{\domain}$ and $\pose^{\domain}$ respectively.

\subsection{Asymmetric Cycle-consistency Learning}\label{sec:cyclegan}
Ideally, in a standard motion-to-motion retargeting approach where target motion data is available, the objective would be to learn a mapping \(\StoT\colon\source\to\target\) that maximizes the likelihood of the output in the distribution of the target domain, \ie, \(p_{\target}\left( \StoT\left( \motion^{\source} \right) \right)\).
In the absence of target motion data, our objective changes to learning a mapping \(\StoT:\source\to\left[\target_{\fid}\right]_{\fid=1}^{\nframe}\)
\begin{equation}
\argmax_{\StoT} p_{\target}\left( \mpose^{\target}_{\fid} \right)
\,\textrm{\st} \left[ \mpose^{\target}_{\fid} \right]_{\fid=1}^{\nframe} = \StoT\left( \motion^{\source} \right).\label{eq:objective}
\end{equation}
In other words, we aim to transfer the source motion such that each frame in the output adheres to the pose priors observed in the target domain's pose data.
Since the data in the source and target are unpaired, \ie pose-level correspondence is absent, we adopt a CycleGAN~\cite{zhu2017unpaired} framework, following the approach of prior works addressing unpaired motion retargeting~\cite{aberman2018deep,aberman2020skeleton,gao2018automaticUS}.

As shown in \Cref{fig:overview}, our model forms an \emph{asymmetric} cycle.
The first half maps a given source motion \(\motion^{\source}\) to a set of poses and root transformations (discussed in \cref{sec:cyclegan}) in the target domain:
\begin{equation}
\left( \mposeSeq, \mrootSeq \right)=\StoT\left( \motion^{\source} \right),\label{eq:halfcycle1}
\end{equation} whereas the other half maps the outputs back to motion in the source domain:
\begin{equation}
\mmotion^{\source} = \TtoS\left( \mposeSeq, \mrootSeq \right).\label{eq:halfcycle2}
\end{equation}
Correspondingly, a pose discriminator \(\D_{\pose}\) and a motion discriminator \(\D_{\motion}\) distinguish the outputs of the two half cycles against real pose and motion samples, respectively.

Following prior work~\cite{aberman2020skeleton}, we supervise the cycle with Wasserstein adversarial loss with gradient penalty~\cite{gulrajani2017improved} and a reconstruction loss.
In summary, the cycle-consistency loss in our model can be written as
\begin{align}
\Loss{cycle} =& \Loss{GAN}\left( \StoT, \D_{\pose} \right) + \lambda_{\textrm{GP}}\Loss{GP}\left( \D_{\pose} \right) +\nonumber & \textrm{first cycle}\\
 & \Loss{GAN}\left( \TtoS\circ\StoT, \D_{\motion}\right) + \lambda_{\textrm{GP}}\Loss{GP} \left( \D_{\motion} \right) + \nonumber & \textrm{second cycle}\\
 & \lambda_{\textrm{recon}}\Loss{recon}\left( \StoT, \TtoS \right),\label{eq:loss_cycle}
\end{align}
where \(\Loss{GAN}\) and \(\Loss{GP}\) are the standard Wasserstein adversarial loss and gradient penalty 
, and \(\Loss{recon}\) is the reconstruction loss
defined as the L2 distance between the input source motion and the remapped source motion 
\begin{align}
\Loss{recon} = & \left\|\mmotion^{\source} - \motion^{\source}\right\|^{2}_{2}. \label{eq:loss_recon}
\end{align}


Our network inherits the architecture designs from prior work~\cite{aberman2020skeleton,SMPL-X:2019,li2022ganimator}.
The mapping networks $\StoT$ and $\TtoS$ is constructed of multiple layers of skeleton-aware operators to effectively account for different joint hierarchies \cite{aberman2020skeleton}.
The motion discriminator, $\D_{\motion}$, also uses skeleton-aware operators and adopts a patch-wise classification to reduce overfitting \cite{li2022ganimator}.
The pose-level discriminator, $\D_{\pose}$, consists of $J+1$ discriminators, one for each joint rotation and another one for all rotations~\cite{SMPL-X:2019}.

\subsection{Root Transformation}\label{sec:root_transformation}
While the above asymmetric CycleGAN framework can generate reasonable motions combining the rough trajectory from the source motion and the pose prior from the target, the generations tend to suffer from a variety of artifacts, including foot sliding and jittering.
The reason is that the root transformations in the target domain has been neglected in the objective defined in \cref{eq:objective}, leading to unresolved ambiguity when mapping the root transformation.
In fact, one trivial (but wrong) solution is to have an identity mapping from the source root motion to the target root motion with scaling, yet this solution leads to various artifacts as shown \cref{sec:zoo}, due to the negligence of changes in the bone size, skeleton structure, \etc.

To address this issue, we propose to \begin{inparaenum}\item[a.)] predict root transformations \(\mrootSeq\) for the target domain, as mentioned in \cref{eq:halfcycle1,eq:halfcycle2}, and \item[b.)] employ a set of soft constraints, described below, to effectively regulate these predicted roots and alleviate the root ambiguity issue. \end{inparaenum}
Although these soft constraints do not directly supervise the root transformations, they promote consistency between the generated and source motion from various complementary perspectives. This guidance helps the root predictions converge towards more realistic and plausible solutions.

\paragraph*{Contact Consistency.}
This constraint focuses on matching the contact-to-ground pattern between the input source motion and the retargeted motion \cite{li2022ganimator}.
In a more intuitive sense, we determine contacts by examining the velocity of the feet joints.
The hypothesis is that when the velocity is close to zero, it indicates a contact point. Therefore, we enforce the retargeted motion to maintain contact whenever the source motion does.
Specifically, using the shorthand \(\motion_{\fid}\) for the motion at frame \(\fid\), we can write the velocity of a specific joint as
\(v_{j}\left( \motion_{\fid} \right) = FK_{j}(\motion_{\fid})-FK_{j}(\motion_{\fid-1})\), where \(FK\) denotes the forward kinematics function that converts joint angles into joint positions $x\in \R^{3J}$.
Then, one can express this constraint using the loss
\begin{equation}
\begin{gathered}
\Ls_{\text{con}} = \dfrac{1}{\nframe\left|\footJ\right|}\sum_{j\in{\footJ}}\sum_{\fid=1}^{\nframe}\left\|v_j(\mmotion^{\target}_{\fid})\right\|^2_2 s_{j}(\motion_{n}^{\source}) \;\text{with}\\
s_{j}(\motion_{n}^{\source}) = \mathbb{1}\left[\left\|v\left(\motion_{\fid}^{\source}\right)\right\|_2<\epsilon\right],
\end{gathered}
\label{eq:contact_label}
\end{equation}
where $\footJ$ represents the set of foot joints and $s_j(\motion_{\fid}^{\source})$ is the reference contact label from source motion, and \(\epsilon\) is the velocity threshold to define contact.

\paragraph*{End-Effectors Consistency.}
End-effectors are the terminal points of a skeleton structure that are commonly used to interact with the real world. 
End-effector consistency takes advantage of the fact that homeomorphic skeletons share a common set of end-effectors, 
and encourages that their normalized velocities from the source and retargeted motions are consistent.
Enforcing this constraint helps prevent common retargeting artifacts like foot sliding~\cite{aberman2020skeleton}. Formally, this constraint is formulated using the following loss
\begin{equation}
    \Loss{ee} = \mathbb{E}_{\MS\sim \PS}\frac{1}{|\eeJ|}\sum_{j\in\eeJ}\left\|\frac{v_{j}\left( \mmotion^{\target} \right)}{h_{Tj}}-\frac{v_{j}(\MS)}{h_{Sj}}\right\|_2^2.
\end{equation}
Here, \(\eeJ\) denote the end-effector joints, and $h^{\source}_j$ and $h^{\target}_j$ correspond to the lengths of the kinematic chains from the root to the end-effector $j$ in the source and target domain, respectively.

Furthermore, under the assumption that the rest poses \(\restP\) in the source and target domain are similar, we require the end-effectors of the source and retargeted motion at every frame \(\fid\) to exhibit comparable offsets to their rest poses.
This objective is based on the premise that if one character's end-effector has moved in a specific direction (relative to its rest pose), the retargeted character should have its corresponding end-effector positioned similarly.
We compute the offsets and this relative end-effector loss using:
\begin{align}
    & o(\motion_{\fid}) = FK(\motion_{\fid})-\restP \\
    &\Loss{ee,r} = \mathbb{E}_{\MS\sim \PS}\dfrac{1}{|\eeJ|}\sum_{j\in\eeJ}\left\|\frac{o_{j}\left(\mmotion_{\fid}^{\target}\right)}{h^{\target}_{j}}-\frac{o_j(\MS)}{h^{\source}_{j}}\right\|_2^2.
\end{align}

In summary, our overall learning objective is
\begin{align}
    \Ls = &\lambda_{\text{cycle}} \Ls_{\text{cycle}} + \lambda_{\text{con}}\Ls_{\text{con}} + \lambda_{\text{ee}}\Ls_{\text{ee}} + \lambda_{\text{ee,r}}\Ls_{\text{ee,r}} .
\end{align}
\vspace{5pt}




\section{Experiments and Evaluations}\label{sec:experiments}

\begin{table*}[tbhp]
    \centering
    \begin{tabular}{ccccccc}
    \toprule
    method & J. Angle Err. \(\downarrow\) & Root Rel J. Pos. Err. \(\downarrow\) & Global J. Pos. Err. \(\downarrow\) & Mean J. Pos. Jitter ($\times 10^2$) \(\downarrow\) & Contact Consis \(\uparrow\) \\\midrule
    \makecell{frame-level SA-Net} &  7.12 &  \textbf{0.53} & 3.88 & 0.81 & 86.6\%\\
    \makecell{motion-level SA-Net} & 14.82 & 1.42 & 1.75 & \textbf{0.47} & 81.2\% \\
    \makecell{\ours{} (ours)} & \textbf{6.73} & 0.54 & \textbf{0.81} & 0.48 & \textbf{91.3}\%\\
    \bottomrule
    \end{tabular}
    \caption{\textbf{Quantitative evaluation on Mixamo.} We compare our approach with the frame-level and motion-level Skeleton-Aware Network~\cite{aberman2020skeleton}, which perform motion retargeting on frame-by-frame and sequence-by-sequence basis, respectively. Our method leverages pose information and further uses the proposed root estimation techniques to achieve more accurate global joint position and higher-quality motion with less jittering and more consistent ground contact.}
    \label{tab:quantitative_mixamo}
\end{table*}

We evaluate our approach using three distinct datasets. 
First, we utilize the Mixamo dataset~\cite{mixamo}, which is a large-scale paired character--motion dataset. Since this dataset provides paired data, it allows us to evaluate the retargeting motions against ground truth for quantitative assessment.
Second, we employ an animal dataset, using a large-scale dog MoCap data from \cite{zhang2018mode} as the source domain, along with a smaller animated animal dataset from \cite{truebones}, containing approximately 1000 frames, as the target domain. This particular setup serves as a stress test to evaluate how our model handles scenarios where the target domains have limited data and differ significantly from the source domain.
Lastly, we extract 3D poses from a horse image dataset \cite{wu2023magicpony} as the target domain and use the dog MoCap dataset as the source domain. This experiment demonstrates our method's capability to learn from accessible but noisy data (in this case, extracted 3D poses obtained from images). To evaluate our results qualitatively, please refer to the supplementary video.

\subsection{Mixamo dataset}
\label{sec:mixamo}

We conduct a quantitative evaluation on the Mixamo dataset \cite{mixamo}, which consists of characters with unique skeletal structures, each performing the same set of 2,400 motion clips. For the source domain, we employ $80\%$ of the motion clips as training samples, with the remaining $20\%$ serving as test samples. For the target domain, we remove temporal information to construct a pose dataset, again splitting it into $80\%$ training samples and $20\%$ test samples. For our quantitative evaluation, we select two distinct characters from the Mixamo dataset (Aj and Mousey), each possessing five primary limbs (two hands, two feet, and a head). Since the Mixamo dataset provides paired motions, we can perform a quantitative evaluation of retargeting performance in comparison to the ground truth.
\subsubsection{Baselines}
We compare our method with two variations of Skeleton-Aware Network~\cite{aberman2020skeleton} (abbreviated as SA-Net), which is a motion retargeting method constructed using the same base operator as our network.

The first variation is a frame-level SA-Net.
In this setup, each domain consists of a set of poses, namely $\PS$ and $\PT$, and two generators and discriminators are trained for translating poses between these domains.
To ensure fairness, the end-effector loss is also incorporated during training. Note that contact consistency loss is not applicable since we cannot compute velocity from single frames.
During retargeting, frame-by-frame decoding is employed.
For root transformation, we approximate the translation from the source root's velocity after scaling it by the skeleton size, and the rotation is directly copied from the source to the target.\\
The second baseline is motion-level SA-Net, which is the original SA-Net. This framework necessitates motion data from both domains, therefore utilizing more information compared to our setting.
We consider this as an upper bound for our approach to provide insights into how our method performs compared to an optimal setting where motion data from both domains are employed. We train the frame-level SA-Net from scratch, and use the checkpoint shared by the authors for the motion-level SA-Net.

\subsubsection{Quantitative Evaluation}
\Tableref{tab:quantitative_mixamo} presents the quantitative evaluation of the different methods described above. For our evaluation, we employ the following commonly used metrics that are widely utilized for assessing the quality of motion reconstruction.
\begin{compactenum}
\item \textbf{Mean Joint Angle Error}: Calculates the joint angle difference (in degrees, represented as axis angles) between the retargeted and ground truth joint angles.
\item \textbf{Mean Root Relative Joint Position Error}: Calculates the MSE between the local joint positions of the retargeted and ground truth motions after removing the global root translation and rotation. The error is normalized by the skeleton's height and multiplied by 1000.
\item \textbf{Mean Global Joint Position Error}: Measures the MSE of the global joint positions between the retargeted and ground truth motions. The error is normalized by the skeleton's height and multiplied by 1000.
\item \textbf{Mean Joint Position Jitter}: Estimates joint position jitter by computing the third derivative (jerk) of the global joint position. A lower value indicates smoother motion, which is generally more desirable.
\item \textbf{Contact Consistency Score}: Calculates the ratio of consistent contacts made between the source and target domains. A contact is considered consistent if the contact state (contact or no contact) determined by \cref{eq:contact_label} is the same in both the source and retargeted motion. A higher Contact Consistency Score indicates better contact consistency.\\
\end{compactenum}

Our method, \ours{}, compares favorably against both baselines across all metrics.
Frame-level SA-Net performs on par in terms of joint angles and relative joint position (see the first two columns), indicating that the relative joint positions can be sufficiently estimated from pose information. Our method leverages this information and further uses the proposed root estimation techniques to achieve more accurate global joint position and higher-quality motion with less jittering and consistent ground contact (see the last three columns).
While motion-level SA-Net should have benefited from receiving additional motion prior in the target domain, it is noticeably worse than our method in most metrics except for joint jittering.
We hypothesize this is because the SA-Net model was pre-trained on multiple target skeletons, therefore it fits relatively less accurately to a specific source-target pair.

\subsection{Animal dataset}
\label{sec:zoo}

We further assess the robustness and versatility of our method by applying it to the challenging task of retargeting animal motion, specifically from dogs to two drastically different animals, T.rexes and hamsters.
As the source domain, we use a large-scale dog MoCap dataset from \cite{zhang2018mode} consisting of 30 minutes of unstructured dog motion encapsulating various locomotion modes. In contrast, the target domain was comprised of a small number of short motion clips of T.rexes and hamsters from the Turebone dataset~\cite{truebones},
from which we extract individual poses as our training data in the target domain.
This evaluation setup presents a high level of complexity due to the large domain gap between the source and target domains. 

\begin{table}
\centering
\setlength{\tabcolsep}{0pt}
\renewcommand{\arraystretch}{0.1}
\resizebox{\linewidth}{!}{%
\begin{tabular}{c*{6}{C{0.13\linewidth}}}
\toprule
\multirow{2}{*}{method}  & \multicolumn{3}{C{0.39\linewidth}}{Mean J. Pos. Jitter ($\times 10^2$) \(\downarrow\)}  & \multicolumn{3}{C{0.32\linewidth}}{Contact Consis \(\uparrow\)}\\
\cmidrule(lr){2-4}\cmidrule(lr){5-7}
& T.rex & Hamster & Horse & T.rex & Hamster & Horse \\ \midrule
\makecell{frame-level SA-Net}  & 5.37 & 8.87 & 2.19 &  86.2\%  & 83.5\% & 77.2\%\\
\makecell{motion-level SA-Net} & \textbf{0.68} & \textbf{0.46} & - & 89.8\% & \textbf{94.6\%} & -  \\
\makecell{\ours{} (ours)} & 1.08 & 1.33 & \textbf{0.88} & \textbf{92.6\%} & 91.1\% & \textbf{81.4\%}\\
\bottomrule
\end{tabular}%
}
\caption{\textbf{Quantitative evaluation for zoo and horse datasets.} Our method largely outperforms frame-level SA-Net~\cite{aberman2020skeleton} in terms of both joint position jitter and contact consistency. While the motion-level SA-Net \cite{aberman2020skeleton} is capable of generating smooth motion, its qualitative results degrade significantly, as shown in \Cref{fig:motionSA}.}
\label{tab:quantitative_zoo}
\vspace{-10pt}
\end{table}

\subsubsection{Motion Quality}
We use the same baselines as in the Mixamo dataset. Both are retrained on this dataset using the same hyperparameters as in the original implementation when possible.
Since there is no ground truth motion data, we evaluate motion quality metrics: Mean Joint Position Jitter and Contact Consistency Score.

\begin{figure}[htbp]
    \centering
    \includegraphics[width=\linewidth]{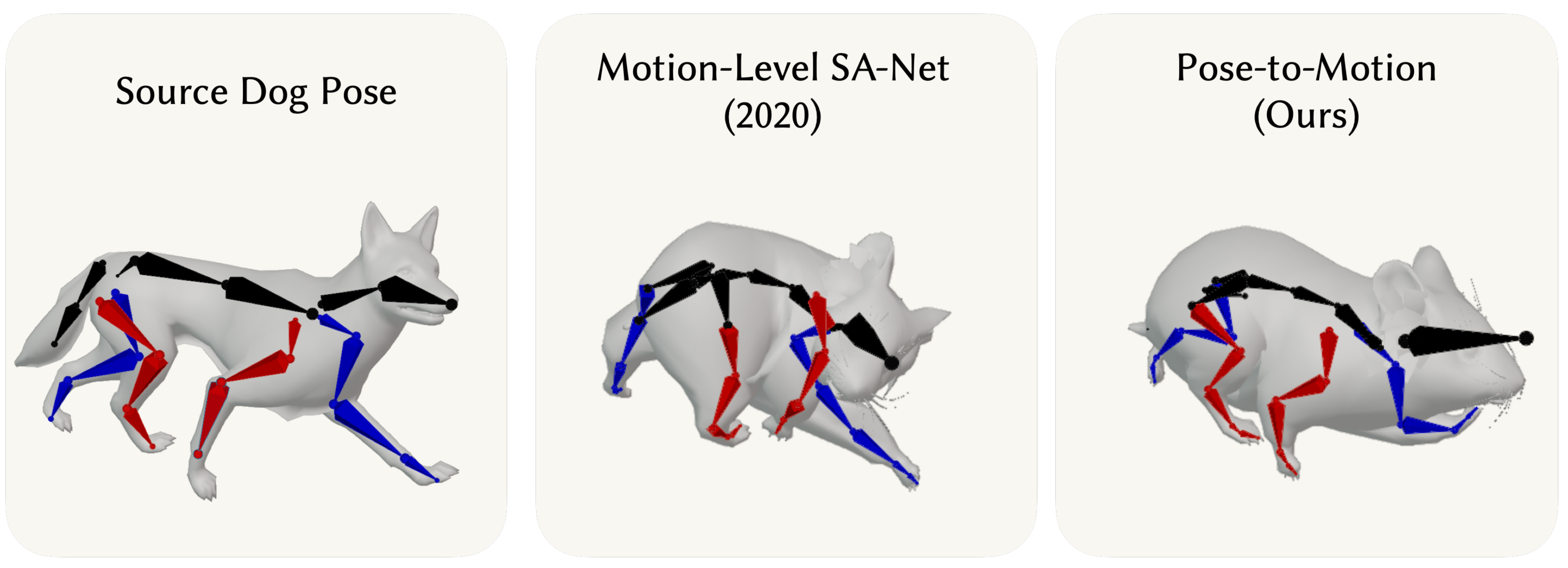}
    \caption{\textbf{Comparison with motion-level SA-Net on the animal dataset.} \ours{} achieves more plausible poses, compare to the motion-level SA-Net, despite the motion data in the target domain being absent. We believe the reason for this performance decline is that motion-to-motion mapping is a much harder task requiring significantly more amount of data and diversity for convergence. For more qualitative evaluations, please watch our supplementary video.}
    \label{fig:motionSA}
\end{figure}

As shown in \Cref{tab:quantitative_zoo} and \Cref{fig:zoo_variety} (T.rex and Hamster), despite the large domain gap and the very limited amount of training poses (600 frames for hamster and 7000 frames for T.rex), our method is able to synthesize high-quality motion. In contrast, the frame-level SA-Net exhibits high jittering and poorer contact consistency.
While the motion-level SA-Net appears smoother motion, it has noticeably less realistic pose as shown in \Cref{fig:motionSA}. For additional qualitative evaluations, we encourage readers to watch our supplementary video.

\begin{figure}
\centering
\includegraphics[width=\linewidth, clip, trim=0 0.5cm 0 1cm]{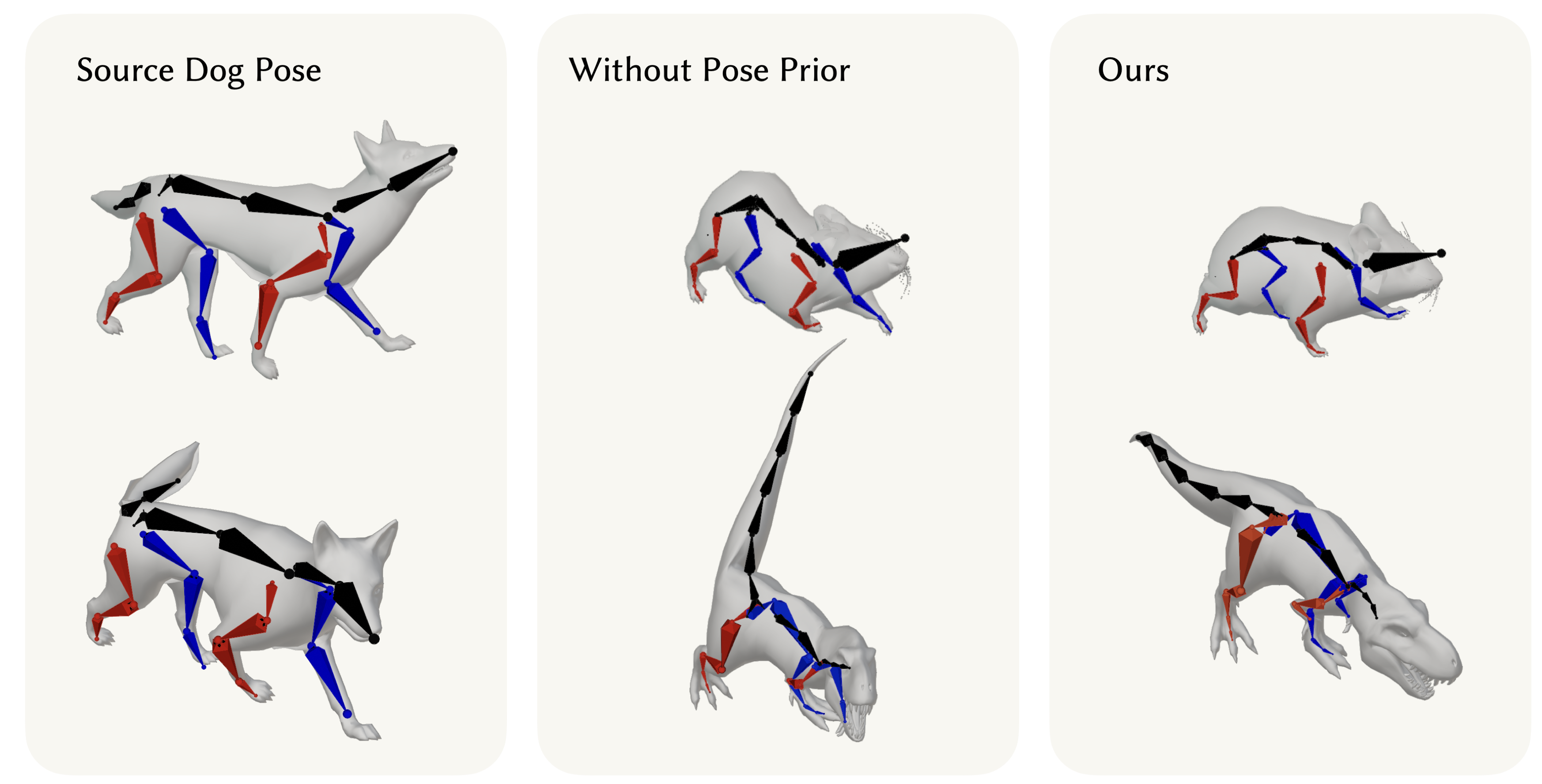}
\caption{\textbf{Preservation of pose characteristics through adversarial training.} We compare the retargeting with and without adversarial training. The latter relies solely on the end-effector and reconstruction loss to establish pose correspondence, thus unable to leverage any pose prior from the target domain, leading to unrealistic and out-of-distribution retargeting results, such as the hamster's head and hip, as well as the T.rex's tail being bent upwards in an unnatural way.}
\label{fig:wo_gan}
\end{figure}

\begin{figure}[htbp]
\centering
\includegraphics[width=\linewidth]{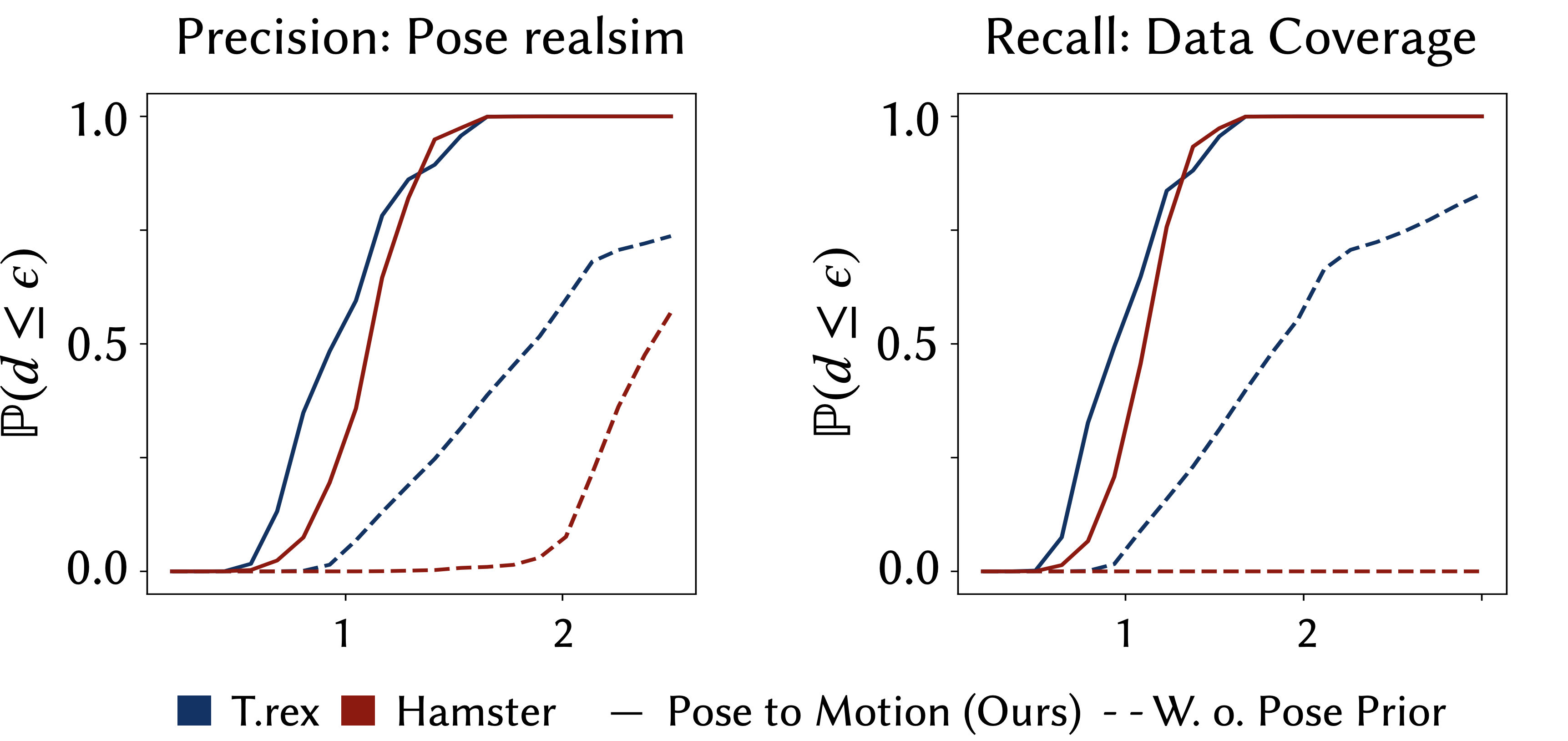}
\vspace{1pt}
    \caption{\textbf{Precision and Recall: }Empirical estimation of data coverage and realism of retargeted poses using precision (left) and recall (right). For both plots, higher values indicate better performance. Across all testing scenarios, our approach (\ours{}) consistently achieved higher values, indicating better pose realism and coverage. Visualizations can be found in \figref{fig:wo_gan}.}
    \label{fig:pr_zoo}
\end{figure}

\subsubsection{Pose Realism and Data Coverage.}
\label{sec:precision_recall}
One of our primary goals is to preserve the diversity and peculiarities of the poses in the target domain.
The retargeted pose should ideally span the entire space of realistic target poses without including extraneous poses.
To evaluate how well our approach achieves this, we employ \textit{Precision} and \textit{Recall} to assess pose realism and data coverage respectively \cite{gan_pose_prior}.
Given \(K\) retargeted poses, precision evaluates the ratio of ``accurate'' predictions.
A retargeted pose is considered accurate if the Mean Root Relative Joint Position Error with at least one sample in the target pose dataset is smaller than a threshold $\epsilon$.
On the other hand, recall measures the ratio of ``covered'' training poses over the size of the training dataset. A training pose is considered covered if the Mean Root Relative Joint Position Error with at least one sample among the retargeted poses is smaller than a threshold \(\epsilon\).

For both, we use \(K=8000\) and plot the precision/recall as a cumulative distribution $\mathbb{P}(d \leq \epsilon)$ in \Cref{fig:pr_zoo}.
Omitting pose prior by removing the adversarial losses (\(\Loss{GAN}\) and \(\Loss{GP}\)) leads to a significant deterioration in precision and recall. This highlights the effectiveness of GAN training in generating realistic and diverse retargeted poses that cover the entire distribution of target poses.
\Cref{fig:wo_gan} visually illustrates this effect. In the absence of adversarial losses, the retargeted poses retain traits from the source domain but appear unnatural in the target domain.

\subsubsection{User Study}
\label{sec:user_study}
\begin{figure}
    \centering
    \includegraphics[width=\linewidth]{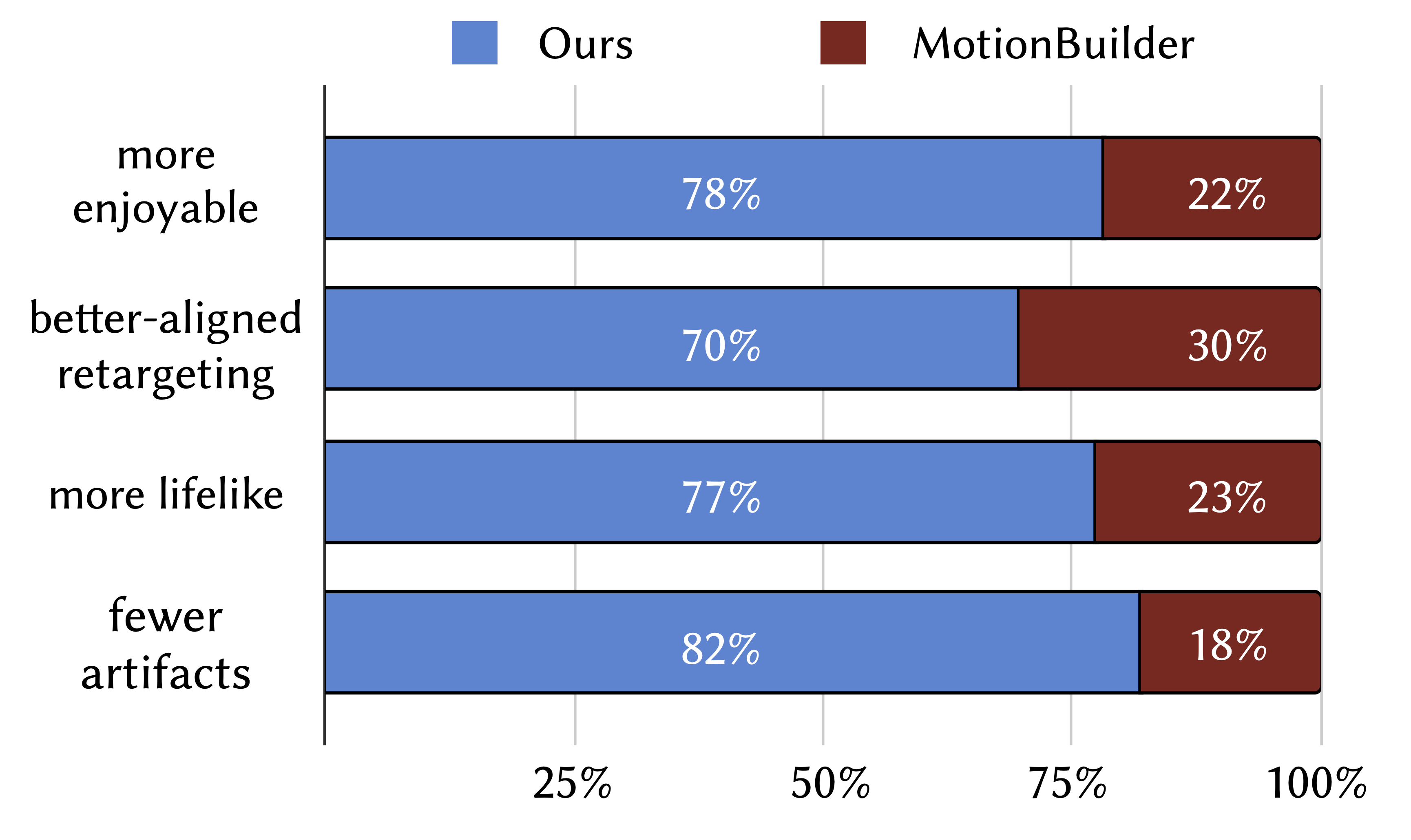}
    \caption{\qq{\textbf{User Study:} We conducted a user study to evaluate the quality of retargeting achieved through our method, comparing it with the commercial software MotionBuilder. In this study, nine motion clips were generated for three distinct characters — Hamster, T.rex, and Horse — with 26 users participating in total. The specific questions posed to the participants during the study are detailed in \cref{sec:user_study}. Generally, the feedback indicated that participants preferred the retargeting results of our method, finding them to be more enjoyable to watch (78\%), noticing that they produced fewer artifacts (82\%), exhibited greater lifelikeness (77\%), and better aligned with the source motion (70\%).}}
    \label{fig:user_study}
\end{figure}

\qq{
We also conduct a user study to evaluate the quality of retargeted motion, where we compare it with commercial motion processing software - MotionBuilder\cite{MotionBuilder}. Note that MotionBuilder requires the manual setting of the correspondences between skeletons with some template skeleton, which is not required in our approach. We rendered 9 motion clips for 3 characters in our experiments - Hamster, T.rex, and Horse. For each motion clip, we ask human subjects the following questions:
\begin{itemize}
    \item {\textbf{Q1:} Which one is more pleasing to watch?}
    \item {\textbf{Q2:} Which adapted animation on the right captures the essence of the original animation displayed on the left side of the video more effectively?}
    \item {\textbf{Q3:} Which adapted animation on the right exhibits better smoothness, lifelikeness, and overall visual appeal?}
    \item {\textbf{Q4:} Which adapted animation on the right shows fewer noticeable issues, such as overlapping body parts or unnatural movement of feet?}
\end{itemize}
Among 26 participants, 78\% found our results more pleasing to watch, 82\% reported observing fewer artifacts, 77\% noted an increase in lifelikeness, and 70\% recognized a closer alignment with the source motion, as depicted in \figref{fig:user_study}. These findings indicate the effectiveness of our approach in generating high-quality motion.}

\subsubsection{Ablation Study}
\begin{figure}
    \centering
    \includegraphics[width=0.7\linewidth]{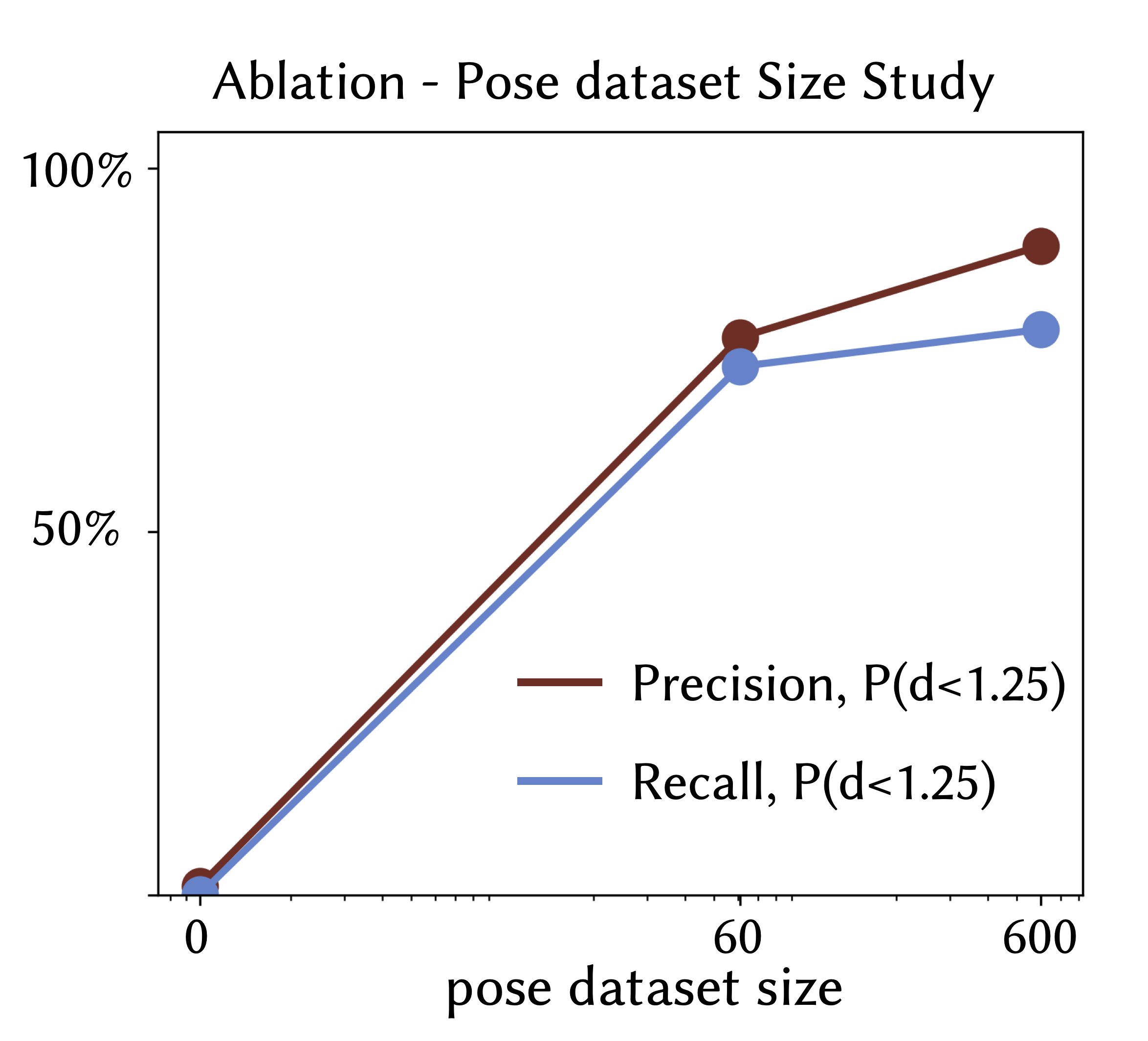}
    \caption{\qq{\textbf{Dataset size ablation:} In this table, we examine the impact of dataset size on the final performance of our approach. We present precision and recall for training sessions with different pose dataset sizes, 0\% (0 poses), 10\% (60 poses), and 100\% (600 poses). We observe that even with 60 poses only, our approach is able to learn meaningful pose features, and yields results with relatively high precision and recall.}}
    \label{fig:ablation}
\end{figure}

\qq{To examine the impact of dataset size on the final performance of our approach, we conducted an ablation study on the Hamster character, comparing the outcomes of training with 0\% data (0 poses), 10\% data (60 poses), and 100\% data (600 poses). Note that the 0\% data scenario corresponds to the case where no pose prior is available; in other words, only the end-effector loss described in \Cref{sec:root_transformation} is utilized during the training process. The evaluation metrics included precision (to measure pose realism) and recall (to measure pose coverage), defined in \Cref{sec:precision_recall}. We observe that even with a smaller dataset, only 60 poses, our method was able to learn meaningful pose features, yielding results that exhibited both high pose realism and coverage, as shown in \Figref{fig:ablation}.}

\subsection{Horse dataset from Images}
\begin{figure}
    \centering
    \includegraphics[width=\linewidth]{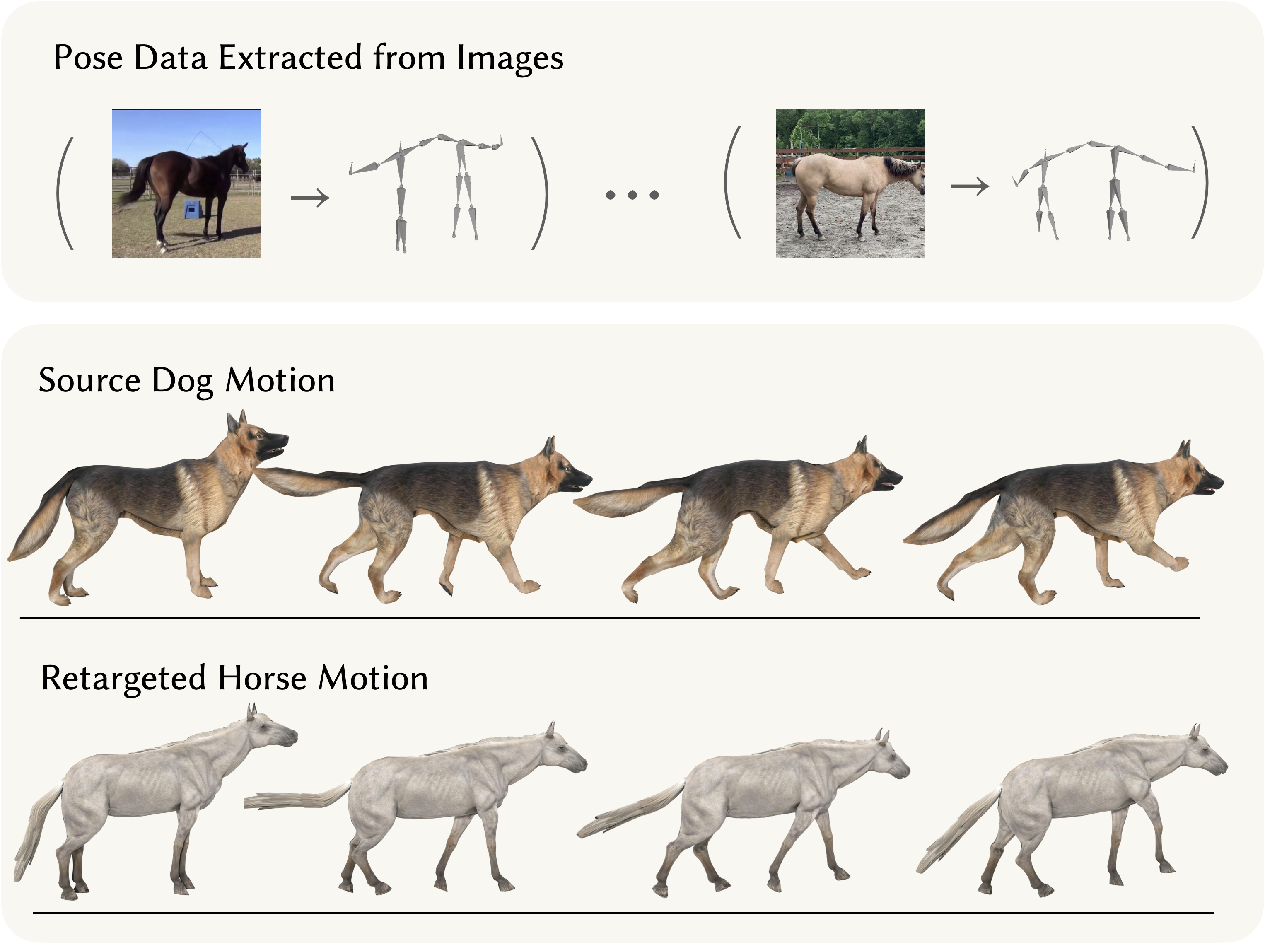}
    \caption{\textbf{Retargeting using noisy pose data estimated from 2D images.} We extract pose priors from a noisy pose dataset generated from state-of-the-art 3D reconstructions method developed in the vision community, and demonstrate that it is possible to synthesize coherent and plausible horse motions by retargeting a dog motion sequence to the horse domain, essentially enabling conditional 2D-to-4D synthesis. 
    }
    \label{fig:horse}
\end{figure}

Our approach is evaluated using pose datasets obtained from images.
To extract 3D poses from a diverse collection of horse images, we utilize the unsupervised method called MagicPony~\cite{wu2023magicpony}.
The dataset used in our evaluation consisted of approximately 10,000 images, capturing various horse poses from different viewing angles.
These extracted 3D poses from MagicPony served as our target domain pose data. We further augmented our dataset by flipping the left and right limbs of the horse.
For the source domain, we use the same dog MoCap dataset described in \Cref{sec:zoo}.
The purpose of this experiment was to demonstrate the robustness and versatility of our approach when applied to readily available but potentially noisy image-derived datasets.
\Cref{fig:horse} visually illustrates the pipeline.
As evident in \cref{fig:horse,fig:zoo_variety}, the retargeted horse poses match the dog poses in the source domain, while at the same time preserving the important features unique to horses, \eg, forward-bending knee, less upright head, and smaller strides.

\label{sec:horse}

\section{Discussion and Conclusion}
Our work tackles the challenging task of synthesizing plausible motion in the absence of reference motion data. We propose a novel approach that leverages static pose data of various characters to generate their motion by projecting the motion prior from another domain with MoCap data and hallucinating plausible root joint movement.
We demonstrate the effectiveness of our approach on a variety of datasets, including the Mixamo dataset, the Animal Pose dataset, and the Horse dataset from images. These experiments show that the proposed method can generate high-quality motion sequences that are both plausible and diverse, and that it can gracefully handle skeletons with significantly different topologies, sizes, and proportions, and even outperforms motion-to-motion retargeting in the low-data regime.

\paragraph*{Limitation.}
While we demonstrated that pose data can provide extremely useful priors for motion synthesis, there are some limitations that inevitably arise from the lack of reference motion data in the target domain. As we transfer the motion prior from the source domain, the generated motion can contain motion traits from the source domain that are unrealistic or physically infeasible for the target domain. For example, dogs have specific gaits that are different from those of horses, our method is not able to account for such differences.
Similarly, the motion prior from the source domain may not be able to capture the full range of motion of the target domain.
One promising venue for future work is to combine pose and limited motion priors to generate more realistic motion.

\paragraph*{Conclusion.}
In this paper, we introduced a neural-based motion synthesis approach through retargeting, leveraging static pose data from the target domain to overcome the restrictive requirement of high-quality motion data.
Our approach opens up new possibilities for motion synthesis in domains where motion data is scarce or unavailable.
By utilizing the latest advancements in related fields such as computer vision, our method can potentially stimulate new applications, such as 2D-to-4D generation, to create new engaging and interactive experiences in entertainment, education, telecommunications and beyond.


\printbibliography                

\end{document}